\documentclass[journal]{IEEEtran}

\ifCLASSINFOpdf
 \else
\fi

\usepackage[dvipsnames]{xcolor}
\usepackage{amsmath}
\usepackage{amssymb}
\usepackage{graphicx}
\usepackage{comment}
\usepackage{xcolor}
\usepackage{tabularray}
\usepackage{adjustbox}
\usepackage{cite}
\usepackage{url}
\usepackage{multirow}
\usepackage{multicol}
\usepackage{comment}
\usepackage{subcaption}
\usepackage{booktabs} 
\usepackage{algorithm}
\usepackage{algpseudocode}
\usepackage{caption}

\begin{document}

\title{Defending Against Knowledge Poisoning Attacks During Retrieval-Augmented Generation}

\author{
    \IEEEauthorblockN{Kennedy Edemacu\IEEEauthorrefmark{1},
    Vinay M. Shashidhar\IEEEauthorrefmark{2}, Micheal Tuape\IEEEauthorrefmark{3}, Dan Abudu\IEEEauthorrefmark{4}, Beakcheol Jang\IEEEauthorrefmark{5}, Jong Wook Kim\IEEEauthorrefmark{6}}

    \thanks{\IEEEauthorrefmark{1}Department of Computer Science, The City University of New York - College of Staten Island, Staten Island, NY, USA. (Email: kennedy.edemacu@csi.cuny.edu). Corresponding Author.}
    \thanks{\IEEEauthorrefmark{2}Department of Mathematics and Computer Science, Northern Michigan University, Marquette, MI, USA. (Email: vmadanbh@nmu.edu).}
    \thanks{\IEEEauthorrefmark{3}Department of Software Engineering, Lappeenranta-Lahti University of Technology, Lappeenranta, Finland. (Email: Micheal.Tuape@lut.fi).}
    \thanks{\IEEEauthorrefmark{5}Energy and Bioproducts Research Institute, Aston University, Birmingham, UK. (Email: d.abudu@aston.ac.uk).}
    \thanks{\IEEEauthorrefmark{4}Graduate School of Information, Yonsei University, Seoul, South Korea. (Email: bjang@yonsei.ac.kr).}
    \thanks{\IEEEauthorrefmark{6}Department of Computer Science, Sangmyung University, Seoul, South Korea. (Email: jkim@smu.ac.kr). Corresponding Author.}
}

\maketitle

\begin{abstract}
Retrieval-Augmented Generation (RAG) has emerged as a powerful approach to boost the capabilities of large language models (LLMs) by incorporating external, up-to-date knowledge sources. However, this introduces a potential vulnerability to knowledge poisoning attacks, where attackers can compromise the knowledge source to mislead the generation model. One such attack is the PoisonedRAG in which the injected adversarial texts steer the model to generate an attacker chosen response for a target question. In this work, we propose novel defense methods, FilterRAG and ML-FilterRAG, to mitigate the PoisonedRAG attack. First, we propose a new property to uncover distinct properties to differentiate between adversarial and clean texts in the knowledge data source. Next, we employ this property to filter out adversarial texts from clean ones in the design of our proposed approaches. Evaluation of these methods using benchmark datasets demonstrate their effectiveness, with performances close to those of the original RAG systems. 

\end{abstract}


\IEEEpeerreviewmaketitle

\section{Introduction}
\IEEEPARstart{A} {key} challenge associated with large language models (LLMs) \cite{brown2020language, achiam2023gpt, anil2023palm} is their tendency of becoming outdated and struggling to integrate the most recent knowledge \cite{tan2024knowledge, zou2024poisonedrag}. This fundamental short-coming is addressed by the recent emergency of retrieval-augmented generation (RAG) \cite{lewis2020retrieval, karpukhin2020dense, borgeaud2022improving, thoppilan2022lamda}. Typically, a RAG system comprises two phases: \textit{Retrieval} and \textit{Generation}. The retrieval phase is accomplished through two components: a \textit{retriever} and a \textit{knowledge database}, while generation is performed by an LLM. During retrieval, the retriever retrieves information relevant to a user's query from the knowledge database. The retrieved text is then passed as a context to the LLM together with the user query to generate the final answer. Various studies have demonstrated the effectiveness of RAGs for different real-world applications \cite{semnani2023wikichat, lozano2023clinfo, shinn2023reflexion, yao2023react}. 

RAG has attracted significant attention from the research community in recent years, with a primary focus on improving its effectiveness and efficiency \cite{asai2023self, izacard2021unsupervised, xiong2020approximate, peng2025soft, kassner2020bert}. However, more recently, an emerging body of research, although limited, has begun to explore the security aspects of RAG systems \cite{zou2024poisonedrag, gong2025topic, nazary2025stealthy, tan2024knowledge, wei2024instructrag, xiang2024certifiably}. Specifically, \cite{zou2024poisonedrag, gong2025topic, nazary2025stealthy} have developed knowledge poisoning attacks for RAG, with PoisonedRAG \cite{zou2024poisonedrag} standing out as an initial significant contributor towards this direction.

In PoisonedRAG \cite{zou2024poisonedrag}, an attacker compromises the knowledge database by injecting adversarial texts. The primary objective is to manipulate the RAG system to generate an attacker-desired response for a specific target question, particularly when using the retrieved, poisoned context during the generation phase. As an example, for a user query, \textit{Which disease is normally caused by the human immunodeficiency
virus?} The attacker can compromise the knowledge database so that the text retrieved from the compromised knowledge database will drive the generation LLM to output \textit{Syphilis} instead of \textit{AIDS} as the final RAG response. Such attacks are highly practical and pose a significant threat, particularly within fact-sensitive application areas such as healthcare, finance, and scientific research, where their impact can be detrimental.

Existing defenses against knowledge poisoning attacks in RAG systems remain limited. Some approaches attempt to detect adversarial texts by analyzing internal LLM activations~\cite{tan2024knowledge}. While this is a promising direction, such methods require white-box access to the generation model. In practice, many modern LLMs are  proprietary and provide only black-box access, which limits the applicability of these approaches. Other approaches rely on ensemble-based generation strategies that combine responses from multiple language models~\cite{xiang2024certifiably}. However, these methods can remain vulnerable when an attacker injects a sufficiently 
large number of adversarial texts into the knowledge database, allowing the poisoned texts to dominate the retrieved context.

In this work, we propose a filtration-based defense mechanism designed to mitigate knowledge poisoning attacks in RAG systems. Our method introduces an additional filtration phase between the retrieval and generation stages, with the goal of 
identifying and removing adversarial texts from the retrieved context before they are passed to the generation model. Unlike prior defenses, our approach operates in the black-box setting and does not require access to model internals. Instead, we leverage statistical properties of the retrieved texts to differentiate adversarial texts from clean texts.

To realize this idea, we develop two approaches: \textit{FilterRAG} and \textit{ML-FilterRAG}. FilterRAG is a threshold-based approach that leverages a new statistical property referred to as \textit{frequency density} (Freq-Density), which measures the concentration of query–answer related words within a retrieved text. This property captures the tendency of adversarial passages to contain a dense overlap of words related to both the query and the attacker-desired answer. By applying an appropriate threshold on this property, adversarial texts can be identified and filtered out before the generation phase.

One potential challenge with FilterRAG is the selection of an appropriate threshold value. To address this challenge, we introduce ML-FilterRAG, which incorporates multiple statistical features and employs a lightweight machine learning model to classify retrieved texts as adversarial or clean. We evaluate the effectiveness of our proposed methods using benchmark datasets and observe that they achieve strong 
defense performance while maintaining accuracy close to that of the original RAG systems. We summarize our main contributions as follows:

\begin{itemize}
    \item We discover a distinct concentration of query-answer pair words, differentiating adversarial texts from clean texts in PoisonedRAG, and propose a new property \textit{Freq-Density} to quantify these concentrations.
    \item We propose two methods, FilterRAG and ML-FilterRAG, to filter out adversarial texts from the retrieved texts. 
    \item We evaluate the effectiveness of these methods against established baselines, observing that their performance levels are comparable.
\end{itemize}
The rest of the paper is organized as follows. Section \ref{sec:preliminaries} presents the preliminaries. In Section \ref{sec:method} we present the methodology. Sections \ref{sec:experiment} and \ref{sec:results} present the experimental setup and the results. Section \ref{sec:conclusion} concludes the paper.

\section{Preliminaries}\label{sec:preliminaries}
\subsection{Threat Model}
\textbf{Attacker's Goal:} We maintain the same threat model as PoisonedRAG \cite{zou2024poisonedrag}. We assume that the attacker pre-selects a set of target questions, denoted as $\mathcal{Q}=\{q_1, q_2,\cdots, q_m\}$, and a corresponding set of desired responses $\mathcal{R}=\{r_1, r_2,\cdots, r_m\}$. The attacker's goal is to subvert the RAG system so that for each target question $q_i\in \mathcal{Q}$, the system generates the attacker-desired response, $r_i\in \mathcal{R}$. 
For example, if the target question is, \textit{Which disease is normally caused by the human immunodeficiency virus?} the compromised RAG system would erroneously answer \textit{Syphilis} instead of the correct response \textit{AIDS}. 

\textbf{Attacker's Capabilities:} For each target question $q_i$, we assume that the attacker can inject a set of $n$ poisoned texts denoted as $\mathcal{P}=\{p_{i}^{1}, p_{i}^{2}, \cdots, p_{i}^{n}\}$ directly into the knowledge database $\mathcal{D}$. We then consider both black-box and white-box settings of the RAG retriever. In the black-box setting, we assume that the attacker has no access to the retriever's parameters and cannot directly query it. However, in the white-box setting, the attacker is assumed to have access to the retriever's parameters. The white-box setting is vital especially considering the potential use of publicly available retrievers such as WhereIsAI/UAE-Large-V1 \cite{li2023angle}, and because it allows evaluations adhering to the well-established Kerchoffs'principle \cite{petitcolas1883cryptographie}. 

\subsection{The Knowledge Corruption Attack}
The knowledge corruption attack in PoisonedRAG is a constrained problem, in which a set of malicious texts $\tilde{\mathcal{D}} = \{p_{i}^{j}|i=1,2,\cdots,m, j=1,2,\cdots,n\}$ is constructed such that the LLM in the RAG system generates the attacker-desired response $r_i \in \mathcal{R}$ for the target question $q_i \in \mathcal{Q}$ when utilizing $k$ texts retrieved from the compromised knowledge database $\mathcal{D}\cup \tilde{\mathcal{D}}$. 

Formally, the attack is represented as: 
\begin{equation}\label{eq:poisonedragopt}
    \max_{\mathcal{\tilde{D}}}\frac{1}{m}\sum_{i=1}^{m} \mathbb{I}\left(LLM\left(q_i;\mathcal{E}(q_i;\mathcal{D}\cup \mathcal{\tilde{D}})\right)= r_i\right)
\end{equation}
where; $\mathbb{I}(.)$ is an indicator function with value 1 if the condition is satisfied and 0 otherwise. $\mathcal{E}(q_i;\mathcal{D}\cup \mathcal{\tilde{D}})$ is the $k$ texts retrieved from the compromised knowledge database $\mathcal{D}\cup \mathcal{\tilde{D}}$.

\subsection{Motivation}
\begin{figure}[t!]
    \centering
    \includegraphics[width=\columnwidth]{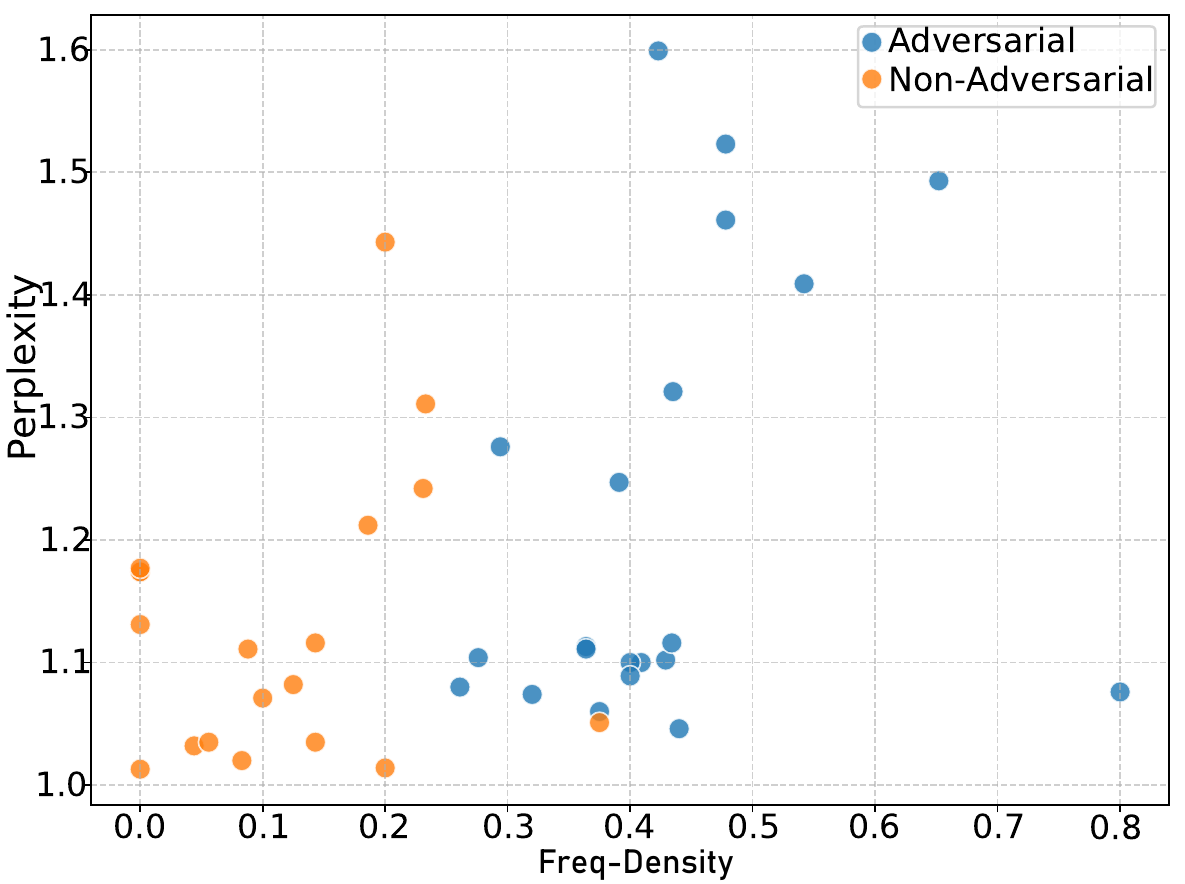} 
    \caption{Pair Plot for Freq-Density vs Perplexity} 
    \label{fig:freq-ratio-perplexity}
\end{figure}

\begin{figure*}[htp]
    \centering
    \includegraphics[width=\textwidth]{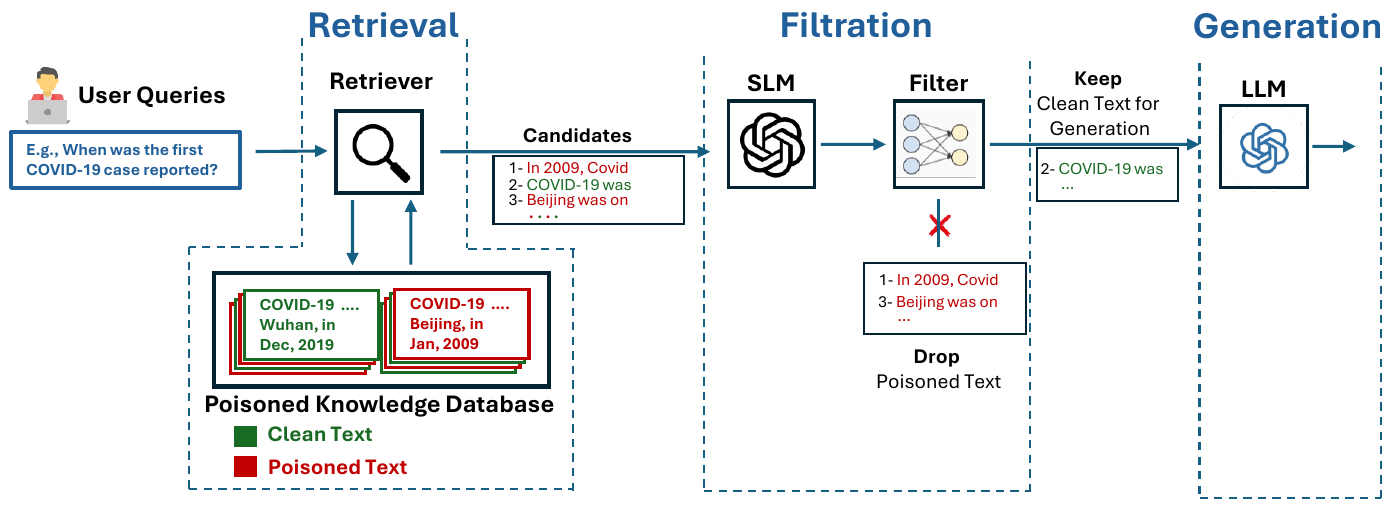} 
    \caption{A high-level illustration of our framework. A filtration phase is integrated into the tradition RAG components to filter-out adversarial texts before the generation phase.} 
    \label{fig:proposed_design}
\end{figure*}

Equation \ref{eq:poisonedragopt} shows that a necessary condition for the success of the knowledge corruption attack is that the malicious text $\mathcal{P}$ appears in the top-$k$ retrieved texts for the target question $q_i$, i.e.,
$\mathcal{P}\in \mathcal{E}(q_i;\mathcal{D}\cup \mathcal{\tilde{D}})$. To achieve this objective, PoisonedRAG constructs adversarial texts that both rank highly during the retrieval phase and steer the RAG LLM toward the attacker-desired response $r_i$ during the generation phase. In order
to satisfy these objectives, adversarial texts typically contain words that are strongly related to both the target query and the attacker-desired answer. As a result, such texts often exhibit a dense overlap of query-related and answer-related words.

In contrast, clean texts retrieved from the knowledge database generally focus on broader contextual information and do not contain such concentrated overlaps. This observation suggests that adversarial texts may exhibit
distinct statistical properties compared to legitimate texts. Statistical properties have previously been used to detect malicious text samples in NLP. For example, \cite{mozes2020frequency} introduced a rule-based method
that identifies adversarial texts by analyzing discrepancies in word frequencies between adversarial and original texts.

Motivated by these observations, we introduce a statistical property referred to as \textit{frequency density} (Freq-Density), which measures the concentration of query--answer related words within a retrieved text. By quantifying how densely such words appear in a text, the Freq-Density property provides a useful signal for distinguishing adversarial texts from clean texts within the retrieved candidates. When combined with other features, it can also provide useful insights into the structure of the retrieved data. For example, Figure~\ref{fig:freq-ratio-perplexity} visualizes Freq-Density against perplexity (see Appendix~\ref{app:pairplots} for additional
visualizations), where distinct clusters of clean and adversarial data points can be observed, supporting our hypothesis.

\section{Methodology} \label{sec:method}

\subsection{Overview}

In this study, we introduce methods to defend against the PoisonedRAG knowledge attack.  Our approaches are based on filtering adversarial texts from the retrieved texts of a 
poisoned knowledge database. Figure~\ref{fig:proposed_design} presents a high-level  illustration of our method. Our framework introduces an additional phase,  \textit{filtration}, that augments the traditional retrieval and generation phases of RAG systems. While the retrieval and generation phases operate as in standard RAG pipelines, the key distinction is that the retrieved texts may contain both 
clean and adversarial passages.

During the filtration phase, we analyze the retrieved texts and remove those that exhibit adversarial characteristics before passing the remaining texts to the generation phase. To accomplish this, we employ a smaller language model (SLM) 
that approximates the generative behavior of the LLM while maintaining lightweight computation. In particular, given a query $q_i$ and a retrieved text $d_j$, the SLM generates an intermediate output $a_j$ that reflects the response the generation model may produce when conditioned on both the query and the retrieved text. This intermediate output provides useful signals about the relationship between 
the query, the retrieved text, and the potential answer.

By combining the query $q_i$ with the generated output $a_j$, we capture query--answer related words that may also appear in the retrieved text $d_j$. This enables us to measure the concentration of such words when computing the Freq-Density property. In contrast, using the query alone would capture only query-related overlap and would not reflect the influence of the retrieved text on the generated answer. The extracted statistical properties are then passed through a filter to identify and remove adversarial texts. The remaining texts, now free of adversarial influence, are combined and used as context during the generation phase. For practical deployment, the filtration phase can be performed 
on the retrieval side. We present the detailed design of this filtration phase in the subsequent sections.

\subsection{Filtration of Adversarial Texts}
In this section, we introduce our proposed frameworks for filtering out adversarial texts. The frameworks aim to minimize the influence/presence of adversarial texts within the context used during the generation phase.

\textbf{Problem Statement:} Given a set of target queries $\mathcal{Q} = \{q_1, q_2, \cdots, q_m\}$, a corresponding set of desired responses $\mathcal{R}= \{r_1, r_2, \cdots, r_m\}$ and a poisoned knowledge database $\mathcal{D}\cup \mathcal{\tilde{D}}$, we seek to minimize Equation \ref{eq:poisonedragopt} so that the target query $q_i \in \mathcal{Q}$ does not result in its corresponding desired response $r_i \in \mathcal{R}$. Formally represented as: 

\begin{equation}\label{eq:poisonedragfilter}
\min_{\mathcal{\tilde{D}}} \quad \frac{1}{m}\sum_{i=1}^{m} \mathbb{I}\left( \text{LLM}\left(q_i;\mathcal{E}(q_i;\mathcal{D}\cup \mathcal{\tilde{D}})\right) = r_i \right)
\end{equation}
$\text{subject to} \quad ||\mathcal{\tilde{D}}|| \to 0.$
By incorporating the latter constraint, we reduce the chances of obtaining the desired response $r_i$ for the target query $q_i$. Our proposed methods are:

\subsubsection{FilterRAG (Threshold-Based Approach)}\label{sec:filterfn} This is a two-stage approach based on selecting a threshold value for a statistical property. We summarize the steps in Algorithm \ref{alg:filterrag}.

\begin{algorithm}
\caption{FilterRAG (Threshold-Based Approach)}
\label{alg:filterrag}
\begin{algorithmic}[1]

\State \textbf{Input:} Target Query: $q_i$, Poisoned Knowledge Database: $\mathcal{D} \cup \tilde{\mathcal{D}}$, Integer: top-$\text{s}$, Float: $\epsilon$, SLM, LLM, Retriever
\State \textbf{Output:} List of Clean Context Items (top-$\text{k}$ for LLM prompting)

\State \textbf{Retrieve Candidate Texts:}
\State \quad $RetrievedItems \gets \text{Retriever}(q_i, \mathcal{D} \cup \tilde{\mathcal{D}}, \text{top}-s)$

\State \textbf{Extract Statistical Property (Freq-Density):}
\For{each item $d_j$ in $RetrievedItems$}
    \State $a_j \gets \text{SLM}(q_i, d_j)$
    \State $(q_i \oplus a_j) \gets \text{Concatenate}(q_i, a_j)$
    \State $\text{Freq-Density} \gets \text{Compute}((q_i \oplus a_j), d_j)$ // Compute(.)
    \hspace*{1.5em}computes according to Eq.~\ref{eq:freq-density}
\EndFor

\State \textbf{Filter Adversarial Texts:}
\State $CleanContextItems \gets \text{EmptyList}()$
\For{each $d_j$ in $RetrievedItems$}
    \If{$\text{Freq-Density}[d_j] < \epsilon$}
        \State Add $d_j$ to $CleanContextItems$
    \Else
        \State Discard $d_j$
    \EndIf
\EndFor

\State \textbf{Return Context:}
\State \Return $CleanContextItems$

\end{algorithmic}
\end{algorithm}

\textbf{Stage 1 - Statistical Property Extraction:} Given a query set $\mathcal{Q} = \{q_1, q_2, \cdots, q_m\}$ and a poisoned knowledge database $\mathcal{D}\cup \mathcal{\tilde{D}}$, we leverage a smaller language model (SLM) to extract the \textit{Freq-Density} statistical property for each retrieved item. In particular, during the retrieval phase, for each query $q_i \in \mathcal{Q}$, we retrieve the top-$s$ items from the poisoned knowledge base $\mathcal{D}\cup \mathcal{\tilde{D}}$ rather than the traditional top-$k$ items. Then, for each of the top-$s$ retrieved items, $d_j$ in the top-$s$, we input both the query $q_i$ and $d_j$ into SLM to generate an output $a_j$.
\begin{equation}\label{eq:slm}
    a_j = \text{SLM}(q_i, d_j) \quad \text{for } j \in \{1, 2, \ldots, s\}
\end{equation}
Next, we define our \textit{Freq-Density} as follows:
\begin{equation}\label{eq:freq-density}
\text{Freq-Density} = \frac{\sum_{w \in (q_i \oplus a_j) \cap d_j} \text{Freq}(w, d_j)}{\text{UniqueWords}(d_j)}
\end{equation}
where, $\oplus$ means concatenation, $(q_i \oplus a_j) \cap d_j$ are semantically similar words common to $(q_i \oplus a_j)$ and $d_j$, i.e., word pairs in $(q_i \oplus a_j)$ and $d_j$ whose computed similarities exceed a predetermined threshold, $\text{Freq}(w, d_j)$ denotes the frequency of the word $w$ within the text $d_j$ and $\text{UniqueWords}(d_j)$ denotes the total number of unique words in $d_j$. The rationale behind this metric is rooted in the attackers' strategy. To both steer the LLM to their desired response $r_i$ for the target query $q_i$, and to ensure high retrieval relevance, attackers populate their adversarial texts with words that are statistically similar to the target queries and their corresponding desired responses. This overlap makes \textit{Freq-Density} a valuable indicator for identifying malicious texts. Figure \ref{fig:freq-ratio-perplexity} presents an empirical illustration of this claim.

\textbf{Stage 2 - Removal of Adversarial Texts:}
We then employ the defined \textit{Freq-Density} to identify adversarial texts for removal by setting a threshold $\epsilon$. We define a filter indicator function (Equation \ref{eq:filter_fn}) to determine whether a retrieved text $d_j$ is adversarial.

\begin{equation}\label{eq:filter_fn}
    \text{Filter}(d_j) = \begin{cases} 
        1, & \text{if } \text{Freq-Density} < \epsilon \\
        0, & \text{otherwise}
    \end{cases}
\end{equation}
We feed each item in the top-$s$ through the filter function. Next, we remove the items whose return value from the filter function is 0. From the remaining items, the top-$k$ items which constitutes the context used to prompt the global LLM are chosen. The effectiveness of this method depends on the choice of $\epsilon$. This parameter is crucial for balancing the trade-off between the overall performance and the presence of adversarial texts in the final top-$k$ items. 

\subsubsection{ML-FilterRAG (Machine Learning Approach)}

\begin{algorithm}
\caption{ML-FilterRAG (Machine Learning Approach)}\label{alg:ml_filterrag}
\label{alg:mlfilterrag}
\begin{algorithmic}[1]

\State \textbf{Input:} Target Query: $q_i$, Poisoned Knowledge Database: $\mathcal{D} \cup \tilde{\mathcal{D}}$, Trained Machine Learning Model: $\mathcal{M}$, top-$\text{s}$, SLM, LLM, Retriever
\State \textbf{Output:} List of Clean Context Items (top-$\text{k}$ for prompting LLM)

\State \textbf{Retrieve Candidate Texts:}
\State \quad $RetrievedItems \gets \text{Retriever}(q_i, \mathcal{D} \cup \tilde{\mathcal{D}}, \text{top}-s)$

\State \textbf{Extract Features:}
\For{each item $d_j$ in $RetrievedItems$}
    \State $a_j \gets \text{SLM}(q_i, d_j)$
    \State $features[d_j] \gets \text{Feature}(q_i, d_j)$
\EndFor

\State \textbf{Predict Adversarial Texts:}
\State $CleanContextItems \gets \text{EmptyList}()$
\For{each $d_j$ in $RetrievedItems$}
    \State $Predict \gets \mathcal{M}(features[d_j])$
    \If{$Predict \text{ is "non-adversarial"}$}
        \State Add $d_j$ to $CleanContextItems$
    \Else
        \State Discard $d_j$
    \EndIf
\EndFor

\State \textbf{Return Context:}
\State \Return $CleanContextItems$

\end{algorithmic}
\end{algorithm}

The effectiveness of our first approach hinges on the choice of $\epsilon$. A high value of $\epsilon$ risks allowing adversarial text samples to appear in the final top-$k$ items. Meanwhile, a low value could filter out legitimately clean texts. Striking the right balance is very crucial and yet challenging. Inspired by the work in \cite{zou2024poisonedrag}, which used perplexity to attempt to filter out adversarial texts, we believe that combining multiple features can effectively address the challenge of $\epsilon$ selection. Thus, we propose ML-FilterRAG, a machine learning-based filtering approach. In this method, a filtering machine learning model takes multiple features extracted from each retrieved text $d_j$ as input and predicts whether $d_j$ is an adversarial sample. Similarly, this is a two-stage method. We summarize the steps in Algorithm \ref{alg:ml_filterrag}.

\textbf{Stage 1 - Feature Extraction:} Similar to the extraction method discussed in \ref{sec:filterfn}, we employ an SLM that takes the target query $q_i$ and a retrieved text $d_j$ as input and subsequently outputs $a_j$. However, on top of the \textit{Freq-Density},  we also compute perplexity, joint log probability of the SLM's output $a_j$ and sum of frequencies of semantically similar words between ($q_i\oplus a_j$) and $d_j$. We incorporate these as supplementary features for our filter model.

\textbf{Stage 2 - Adversarial Text Prediction:} Given a light-weight trained machine learning-based filter model $\mathcal{M}$, each of the retrieved items in the the top-$s$ undergoes the feature extraction process discussed previously. Once these feature values are extracted, they are fed into $\mathcal{M}$ (according to Equation \ref{eq:ml_filterrag}) to predict whether the item $d_j$ is an adversarial text.  
\begin{equation}\label{eq:ml_filterrag}
    pred_j = \mathcal{M}(\text{Feature}((q_i, d_j))
\end{equation}
where, $pred_j$ can be adversarial or non-adversarial, and $\text{Feature}(.)$ is a function that extracts features. Finally, the top-$k$ items are chosen from the non-adversarial items as context to prompt the LLM. By generating features that capture semantic similarities, we believe that our approaches offer an expanded scope for identifying adversarial texts. To train $\mathcal{M}$, we use the supervised learning approach in which the labeled datasets of retrieved texts for each target query are annotated as adversarial and non-adversarial. For each training instance, we extract features using the method discussed above. These features capture the statistical cues indicative of adversarial texts. $\mathcal{M}$ is trained to learn the mappings between these features and their corresponding labels. Once trained, $\mathcal{M}$ serves as an effective filter that predicts if $d_j$ is adversarial based on its computed features.

\section{Experiments}\label{sec:experiment}
\subsection{Datasets}
For our experiments, we utilize three benchmark question-answering datasets: \textit{MS-MARCO} \cite{nguyen2640ms}, \textit{Natural Questions (NQ)} \cite{kwiatkowski2019natural} and \textit{HotpotQA} \cite{yang2018hotpotqa}. Each of these original datasets contains a distinct knowledge database. Specifically, NQ contains 2,681,468 texts, HotpotQA contains 5,233,329 texts, and MS-MARCO contains 8,841,823 texts within their respective knowledge databases. In addition, each dataset provides a set of queries and answers. To facilitate the identification of adversarial texts, we leverage the version of these datasets detailed in \cite{zou2024poisonedrag}. In this version, 100 closed-ended questions from each dataset are designated as target questions, each paired with a ground-truth answer and an attacker-desired answer. Furthermore, for each target question, 5 adversarial texts are injected into the original knowledge database. 

To evaluate our ML-FilterRAG approach, we had GPT-4o generate an additional 5 adversarial texts for each target question, mimicking the characteristics of the original five. We carefully examined each of these newly generated texts to ensure that they meet two crucial criteria: they successfully output the attacker-desired response, and they rank highly among the retrieved texts from the knowledge base. Specifically, we selected those that appeared within the top 15 retrieved results. We then randomly selected five adversarial texts, along with an equivalent number of randomly selected original (clean) texts, to train our machine learning-based filter models. We keep the rest for evaluating our proposed framework. For the white-box attack, we adopt the HotFlip method as in \cite{zou2024poisonedrag}, with all the other settings remaining consistent as for the black-box attack. 

\subsection{RAG Settings}
\subsubsection{Retriever}
We employ the Contriever \cite{izacard2021unsupervised} as our retriever. Following \cite{zou2024poisonedrag, lewis2020retrieval}, and unless otherwise stated, we employ the dot product to measure similarities between queries and  texts within the knowledge database. 

\subsubsection{SLM}
We leverage two language models as our SLM: LLaMA-3 \cite{grattafiori2024llama} and LLaMA-2 \cite{touvron2023llama}. For our first filtration approach, we used a default $\epsilon$ value of 0.2. In contrast, for the machine learning-based approach, we defined and trained a dedicated model for each dataset. Specifically, we employ an XGBoost model for the NQ dataset, while Random Forest models were utilized for both the HotpotQA and MS-MARCO datasets. The training details of the models are presented in the Appendix \ref{app:training-details}. For semantic word similarity matching, we employ a huggingface sentence transformer\footnote{https://huggingface.co/sentence-transformers/all-MiniLM-L6-v2} and set a default similarity threshold value of $0.6$ for cosine similarity. We will also study the impact of the similarity threshold and $\epsilon$ in our evaluations. 

\subsubsection{LLM}
For the generation LLM, we consider a family of GPT models, GPT-3.5 \cite{brown2020language}, GPT-4 \cite{achiam2023gpt} and GPT-4o \cite{hurst2024gpt}, and LLaMA-3. We adopt the system prompt discussed in \cite{zou2024poisonedrag} and we present it in Appendix \ref{app:prompt}. We use a temperature value of $0.1$ for all the models. 

\subsection{Metrics}
\subsubsection{Adversarial Text Ratio (ATR)}
We introduce a new performance metric, \textit{adversarial text ratio (ATR)} to quantify the fraction of adversarial texts within the retrieved top-$k$ texts. This metric is particularly relevant given our objective of minimizing Equation \ref{eq:poisonedragfilter}, an objective directly influenced by the number of adversarial texts. Therefore, ATR is crucial for accurately assessing the effectiveness of our proposed methods.

\subsubsection{Attack Success Rate (ASR)}
We also consider ASR as discussed in \cite{zou2024poisonedrag}. ASR quantifies the fraction of target questions whose answers match the attacker-desired answers. Substring matching between the attacker desired answer and the LLM generated answer is used to determine if the two answers are the same. In other words, the two answers do not need to match exactly.

\subsubsection{Accuracy}
Accuracy measures the fraction of target questions whose ground-truth answers match the answers generated by the LLM. In a similar manner, we adopt the substring matching approach to determine a match between the two answers. We expect higher accuracy values for frameworks with minimal number of adversarial texts in the retrieved top-$k$ texts and vice versa. 

\subsection{Baselines}
\subsubsection{PoisonedRAG}
We compare our framework with PoisonedRAG \cite{zou2024poisonedrag}. In doing so, our aim is to assess whether our proposed approaches minimize the influence of adversarial texts injected into the knowledge database through the PoisonedRAG attack. 

\subsubsection{CleanRAG} Similarly, we consider comparing our methods with the traditional RAG with no adversarial texts within its knowledge database. We hope to achieve performance similar or close to the CleanRAG framework. 

\subsection{Other Experimental Parameters}

To evaluate our methods under a poisoned knowledge database setting, we retrieve a balanced set of adversarial and clean texts within the top-$s$ retrieved candidates. This controlled configuration allows us to systematically analyze how well our proposed approaches can identify and filter adversarial texts when both adversarial and legitimate texts appear among the retrieved results. In practice, the retriever does not have prior knowledge of whether a retrieved text is adversarial or clean, and the proportion of adversarial texts can vary depending on the extent of knowledge poisoning in the knowledge database. By using this balanced setup, we ensure that the filtration phase is consistently exposed to adversarial texts while also preserving legitimate retrieved context.

In our experiments, we set the top-$s$ retrieval to $4$ and intentionally retrieve an equal number of adversarial and clean texts. We then select the top-$k$ of $2$ texts for the generation phase. We use a default value of $10$ for $m$, and repeat each experiment $10$ times so that all target questions are examined.

\section{Experimental Results}\label{sec:results}
\subsection{Main Results}
\begin{table*}[htbp]
    \centering
    \caption{Performance Comparison of RAG Frameworks Across Different LLMs for Black-box Attack}
    \label{tab:main-performance-black-box}
    \begin{tabular}{@{}ll c cc cc cc cc@{}} 
        \toprule 
        \textbf{Dataset} & \textbf{Framework} & \textbf{ATR} $\downarrow$& \multicolumn{2}{c}{\textbf{GPT-3.5}} & \multicolumn{2}{c}{\textbf{GPT-4}} & \multicolumn{2}{c}{\textbf{GPT-4o}} & \multicolumn{2}{c}{\textbf{LLaMA-3}}\\
        \cmidrule(lr){4-5} \cmidrule(lr){6-7} \cmidrule(lr){8-9} \cmidrule(lr){10-11}
        & & & \textbf{ASR} $\downarrow$ & \textbf{Accuracy} $\uparrow$& \textbf{ASR} $\downarrow$ & \textbf{Accuracy} $\uparrow$& \textbf{ASR} $\downarrow$ & \textbf{Accuracy} $\uparrow$ &  \textbf{ASR} $\downarrow$ & \textbf{Accuracy} $\uparrow$\\
        \midrule 
        \multirow{4}{*}{HotpotQA} & CleanRAG & \textbf{0.000} & \textbf{0.080} & \textbf{0.913} & \textbf{0.090}& \textbf{0.908}& \textbf{0.090}& \textbf{0.907} & \textbf{0.310} & \textbf{0.665}\\
        & PoisonedRAG & 1.000 & 0.940 & 0.042 & 0.900& 0.000 &0.930 & 0.000 &0.980 & 0.001\\
        & FilterRAG ($\epsilon=0.2$) & \textbf{0.000} & 0.082 & 0.881 & \textbf{0.090}&0.870 &0.091 & 0.820 & 0.380 & 0.536\\
        & ML-FilterRAG & 0.015 & 0.090 & 0.903 &0.091 &0.905 &0.094 & 0.899 & 0.330 & 0.541\\
        
        \midrule 
        \multirow{4}{*}{MS-MARCO} & CleanRAG & \textbf{0.000} & \textbf{0.060} & \textbf{0.859} & \textbf{0.050}& \textbf{0.851} & \textbf{0.050} & \textbf{0.867} & \textbf{0.310} & \textbf{0.667}\\
        & PoisonedRAG & 0.825 & 0.820 & 0.160 & 0.824 & 0.173 & 0.834 & 0.128 & 0.912 &0.082\\
        & FilterRAG ($\epsilon=0.2$) & 0.065 & 0.090 & 0.820 & 0.090 & 0.840 & 0.080 & 0.840 & 0.430 & 0.509\\
        & ML-FilterRAG & 0.045 & \textbf{0.060} & 0.851 & 0.060 & 0.849 & 0.080 & 0.831 & 0.400 & 0.565\\
        
        \midrule 
        \multirow{4}{*}{NQ} & CleanRAG & \textbf{0.000} & \textbf{0.010} & \textbf{0.831} & \textbf{0.020} & \textbf{0.833} & \textbf{0.020} & \textbf{0.831} &\textbf{0.140} & \textbf{0.575}\\
        & PoisonedRAG & 0.980 & 0.880 & 0.108 & 0.870 & 0.119 & 0.864 & 0.118 & 0.980 & 0.100 \\
        & FilterRAG ($\epsilon=0.2$) & 0.010 & 0.030 & 0.756 & 0.030 & 0.818 & 0.030 & 0.773 & 0.190 & 0.480\\
        & ML-FilterRAG & 0.030 & 0.020 & 0.811 & 0.030 & 0.810 & 0.030 & 0.803 & 0.180 & 0.544\\
        \bottomrule 
    \end{tabular}
\end{table*}

To explore whether our proposed methods can filter out adversarial texts from the retrieved texts with no significant negative impacts on performance, we conduct experiments on the three datasets: HotpotQA, MS-MARCO and NQ datasets under white-box and black-box PoisonedRAG attacks. 
The results are presented in Table \ref{tab:main-performance-black-box}.
From the experimental results, we make the following observations. First, our proposed approaches, FilterRAG and ML-FilterRAG consistently achieve higher accuracies in the presence of adversarial texts. For example, FilterRAG can achieve accuracies of 88.1\%, 82.0\%, and 75.6\% with GPT-3.5 on HotpotQA, MS-MARCO, and NQ datasets, respectively. While, ML-FilterRAG achieves accuracies of 90.3\%, 85.1\% and 81.1\% with GPT-3.5 on HotpotQQ, MS-MARCO, and NQ datasets, respectively. 

We also observe that our proposed FilterRAG and ML-FilterRAG methods achieve some of the lowest ASRs in comparison to the baselines. For example, FilterRAG has ASRs of 8.2\%, 9.0\%, and 3.0\% with GPT-3.5 on HotpotQA,  MS-MARCO, and NQ datasets, respectively. Meanwhile, ML-FilterRAG achieves ASRs of 9.00\%, 6.0\%, and 2.0\% on HotpotQA, MS-MARCO, and NQ datasets, respectively. 

Similarly, FilterRAG and ML-FilterRAG achieve some of the lowest ATRs. For example, FilterRAG has ATR of 0.0\%, 6.5\%, and 1.0\% with HotpotQA, MS-MARCO, and NQ datasets, respectively. While ML-FilterRAG has ATR of only 1.5\%, 4.5\%, and 3.0\% with HotpotQA, MS-MARCO, and NQ datasets, respectively. More results for white-box attack settings are shown in Appendix \ref{app:white-box}.

These results highlight several important observations. First, both \textit{FilterRAG} and \textit{ML-FilterRAG} significantly reduce the presence of adversarial texts in the retrieved context compared to the PoisonedRAG baseline. This reduction directly contributes to the lower ASR values observed for our proposed methods, since fewer adversarial passages remain available to influence the generation model. In particular, by filtering texts that exhibit high concentrations of query--answer related words, our methods are able to remove many adversarial passages before the generation phase.

Second, we observe that \textit{ML-FilterRAG} generally achieves better accuracy than \textit{FilterRAG}. This behavior can be explained by the fact that \textit{FilterRAG} relies on a single threshold applied to the Freq-Density
property, which may occasionally filter legitimate texts that contain strong query-related signals. In contrast, \textit{ML-FilterRAG} combines multiple statistical features when identifying adversarial texts, allowing it to more
accurately differentiate between adversarial and clean texts.

Finally, we observe variations in performance across datasets. For example, datasets such as MS-MARCO often contain longer passages that provide richer contextual information, which makes it easier for the proposed methods to
identify abnormal concentrations of query--answer related words. In contrast, datasets with shorter passages or more concise answers can make this distinction less pronounced. Despite these differences, our proposed methods consistently achieve lower ATR and ASR values while maintaining accuracy close to that of the CleanRAG baseline.

Although our methods are able to achieve lower ATRs, they cannot exceed the CleanRAG baseline in terms of accuracy. This is mainly because some clean texts get misclassified as adversarial texts. However, this misclassification is more common with FilterRAG than with ML-FilterRAG. Setting the right $\epsilon$-value is crucial for the success of FilterRAG. However, this is fairly difficult to achieve. It is a trade-off between ATR and accuracy. But both of our proposed methods show significant improvements compared to the PoisonedRAG baseline, and they achieve performance that closely match the original RAG. 

\subsection{Ablation Studies}
We also performed ablation studies and present the results as follows.
\subsubsection{Varying Similarity Threshold}
\begin{table*}[htbp]
    \centering
    \caption{Performances with Varying Similarity Threshold Values for MS-MARCO Dataset using GPT-4}
    \label{tab:similarity_threshold}
    \begin{tabular}{@{}llccccccc@{}} 
        \toprule 
         \multirow{2}{*}{\textbf{Method}}&\multicolumn{6}{c}{\textbf{Similarity Threshold}} \\
         & & 1.0 & 0.9 & 0.8 & 0.7 & 0.6 & 0.5 & 0.4\\
         \midrule 
         \multirow{3}{*}{FilterRAG}& \textbf{ATR} $\downarrow$ & 0.415 & 0.080 & 0.070& 0.065 &  0.065& 0.020 & \textbf{0.000}\\
         & \textbf{ASR} $\downarrow$ & 0.260 & 0.080&  0.090 & 0.090 & 0.090 & 0.070 & \textbf{0.050}\\
         & \textbf{Accuracy} $\uparrow$ & 0.621 & 0.827 & 0.831  & 0.833 & \textbf{0.840} & 0.766 &0.497\\
         \midrule 

         \multirow{3}{*}{ML-FilterRAG}& \textbf{ATR} $\downarrow$ & 0.425 & 0.090&  0.080&0.060& \textbf{0.045} & 0.195 & 0.480\\
         & \textbf{ASR} $\downarrow$ &  0.290 & 0.090 & 0.086 &0.070 & \textbf{0.060} & 0.120 & 0.230\\
         & \textbf{Accuracy} $\uparrow$ &  0.589 & 0.834& 0.834& 0.844 & \textbf{0.849} & 0.793 & 0.538\\        
        \bottomrule 
    \end{tabular}
\end{table*}
The similarity threshold determines the level of exactness between the query-answer combination and the context text words. This exactness is crucial for both the FilterRAG and ML-FilterRAG methods. As a result, we perform experiments that demonstrate how this exactness affects our proposed methods.
We present the results in Table \ref{tab:similarity_threshold}. From the results, we observe that at the similarity threshold of 1.0, FilterRAG and ML-FilterRAG both have high ATR and ASR, and low accuracies. As the similarity threshold decreases from 0.9 to 0.6, both FilterRAG and ML-FilterRAG show decreasing ATR and ASR, while their accuracies increase. Beyond the similarity threshold value of 0.6, FilterRAG shows decreasing ATR, ASR and accuracy values, while ML-FilterRAG shows increasing ATR and ASR values, but decreasing accuracy values. This result demonstrates that, at highest similarity threshold of 1.0, only query-answer combination words that perfectly match the context text words are considered. This level of perfection can allow attackers to evade our proposed methods simply using modified word versions or synonyms and allow more adversarial texts to go undetected by our methods. A reason our proposed methods show high ATR and ASR values, and low accuracy at similarity threshold of 1.0. As we relax the level of exactness by reducing the similarity threshold, the match between query-answer combination and context text words is now measured at semantic level. This is reflected by the decrease in both ATR and ASR values and the increase in accuracy values for similarity thresholds of 0.9 to 0.6. However, further decrease in similarity threshold values beyond 0.6, blurs the boundary between legitimate and adversarial texts and results in poor performance for both FilterRAG and ML-FilterRAG. 

\subsubsection{Effect of SLM}
\begin{table*}[htbp]
    \centering
    \caption{Performances with Varying SLMs for MS-MARCO Dataset with GPT-4}
    \label{tab:slms}
    \begin{tabular}{@{}lcccccc@{}} 
        \toprule 
         \multirow{2}{*}{\textbf{Method}} & \multicolumn{3}{c}{\textbf{LLaMA-2}} & \multicolumn{3}{c}{\textbf{LLaMA-3}}\\
         \cmidrule(lr){2-4} \cmidrule(lr){5-7}
         & \textbf{ATR} $\downarrow$ & \textbf{ASR} $\downarrow$ & \textbf{Accuracy} $\uparrow$ & \textbf{ATR} $\downarrow$& \textbf{ASR} $\downarrow$ & \textbf{Accuracy} $\uparrow$\\
         \midrule 
          FilterRAG &\textbf{0.215} & \textbf{0.15} &0.786 & 0.065 & 0.090 & 0.840\\
         ML-FilterRAG & \textbf{0.215} & 0.16& \textbf{0.791} & \textbf{0.045} & \textbf{0.060} & \textbf{0.849}\\        
        \bottomrule 
    \end{tabular}
\end{table*}
We also investigate the performance of our proposed methods with two SLMs, LLaMA-2 and LLaMA-3. The results are presented in Table \ref{tab:slms}. From the result, we observe that with LLaMA-2 SLM, FilterRAG has a better ASR value compared to ML-FilterRAG. However, ML-FilterRAG has better accuracy. With LLaMA-3 as SLM, we observe that ML-FilterRAG performs better on all metrics. This result generally shows that using a large model as an SLM achieves a better performance. This makes sense as a more extensive SLM possesses the ability to generate precise and clear results based on the provided context, thereby enabling more accurate matching between query-answer combination and the context text. 

\subsubsection{Effect of $\epsilon$-value}
\begin{table}[htbp]
    \centering
    \caption{FilterRAG Performance with respect to $\epsilon$-Values for MS-MARCO Dataset with GPT-4}
    \label{tab:attack_epsilon_performance}
    \begin{tabular}{@{}lccccc@{}} 
        \toprule 
         &\multicolumn{5}{c}{\textbf{$\epsilon$-value}} \\
         & 0.1 & 0.2 & 0.3 & 0.4 & 0.5 \\
         \midrule 
         \textbf{ATR} $\downarrow$ & \textbf{0.005} & 0.065&0.230 & 0.540 & 0.765\\
         \textbf{ASR} $\downarrow$ &  \textbf{0.050} & 0.090& 0.200 & 0.320 & 0.540\\
         \textbf{Accuracy} $\uparrow$ &  0.624 & \textbf{0.840}& 0.764 & 0.462 & 0.280\\
        
        \bottomrule 
    \end{tabular}
\end{table}
The performance of FilterRAG depends on the choice of $\epsilon$. We demonstrate this through experiments. We present the experimental results for the MS-MARCO dataset in Table \ref{tab:attack_epsilon_performance}. From the result, we observe that at the lowest $\epsilon$ value of 0.1, FilterRAG achieves its lowest ATR and ASR, however the accuracy is low, at only 62.4\%. As $\epsilon$-value increases to 0.2, FilterRAG achieves the best performance at an accuracy of 84.0\%. However, its ATR and ASR begin to drop. Further increase in $\epsilon$-values from 2.0 results in more increase in ATR and ASR values with decreasing accuracy. At lower $\epsilon$-values, the majority of adversarial texts from top-$s$ are filtered out. However, some legitimate texts are mistakenly filtered out too. This is why the ATR and ASR are lower, but the accuracy is relatively low as well. The accuracy begins to increase with the increasing $\epsilon$-values because more legitimate texts are being captured in the top-$k$. But, this affects both ATR and ASR, as more more adversarial texts sneak into the top-$k$. This becomes more evident at higher $\epsilon$-values. This result emphasizes that choosing an $\epsilon$-value is a trade-off between minimizing the number of adversarial texts in the retrieved top-$s$ and achieving better performance.

\subsection{Real-World Deployment Considerations}
To support real-world use, FilterRAG and ML-FilterRAG are designed as modular filtration layers that operate between the retrieval and generation stages of standard RAG systems. This setup allows for straightforward integration without altering the retriever or language model components. After retrieving the top-$s$ candidate texts, statistical features are extracted using an SLM, and adversarial texts are filtered before passing the clean context to the generation phase. This added filtration phase introduces some computational overhead, but it remains tractable for batch and near real-time systems, particularly when using small or quantized SLMs. SLM inference and semantic similarity computations are the main contributors to cost, particularly as the number of retrieved documents increases. However, classifier inference using Random Forest or XGBoost is fast and efficient. To further optimize performance in latency-sensitive applications, strategies such as reducing top-$s$ or dynamically adapting it based on query confidence, or caching feature computations can be employed.

To address a key deployment trade-off between accuracy and efficiency, our ablation studies (Tables \ref{tab:similarity_threshold}–\ref{tab:attack_epsilon_performance}) show that tuning hyperparameters such as similarity threshold and $\epsilon$-value can balance detection accuracy against runtime cost. ML-FilterRAG achieves a stronger trade-off, offering higher accuracy with minimal increase in adversarial text ratio or attack success rate, while maintaining operational feasibility. In general, both of our proposed methods are deployable with manageable overhead, offering practical and scalable defenses for strengthening RAG systems against knowledge poisoning attacks.

\section{Other Related Works}\label{sec:related_works}
\subsection{RAG Knowledge Poisoning Attacks}
In other knowledge poisoning attacks, \cite{cho2024typos} \textit{et al}. introduced GARAG. GARAG corrupts the knowledge database through low-level perturbation to original documents. Zhong \textit{et al}. \cite{zhong2023poisoning} proposed PRCAP, which introduces adversarial texts into the knowledge database. The introduced texts are generated by perturbing discrete tokens that enhance their similarity to target queries. Gong \textit{et al}. \cite{gong2025topic} proposed Topic-fliprag attack. Topic-fliprag performs adversarial perturbations on original documents to influence opinions across target queries. Nazary \textit{et al.} \cite{nazary2025stealthy} proposed a stealthy knowledge poisoning attack for RAG-based recommender systems. Their attack performs adversarial perturbations to item descriptions to either promote or demote the item. These methods have further demonstrated the vulnerability of RAGs to knowledge positioning attacks.

\subsection{Defense Against Knowledge Poisoning}

Several works have explored defenses against knowledge poisoning attacks in retrieval-augmented generation systems. Tan \textit{et al}.~\cite{tan2024knowledge} propose detecting adversarial texts by analyzing internal activations of the LLM. While this approach shows promise, it requires white-box access to the generation model, which limits its applicability in practice since many modern LLMs provide
only black-box access.

Xiang \textit{et al}.~\cite{xiang2024certifiably} propose an ensemble-based defense that generates multiple responses using different LLMs and aggregates them to produce the final answer. Although this method operates in the black-box setting, it can remain vulnerable when an attacker injects a
sufficiently large number of adversarial texts into the knowledge database, allowing poisoned texts to dominate the retrieved context.

Zou \textit{et al}.~\cite{zou2024poisonedrag} suggest several mitigation strategies, including paraphrasing, perplexity-based filtering, and duplicate text removal. However, these techniques are not sufficiently robust to reliably detect and
remove adversarial texts from the retrieved context.

Table~\ref{tab:intro_comparison} summarizes the key differences between these existing defenses and our proposed approaches. In contrast to prior methods, our approaches, \textit{FilterRAG} and \textit{ML-FilterRAG}, do not require
access to internal model activations and do not rely on ensemble generation. Instead, our methods leverage statistical properties of the retrieved texts to identify and filter adversarial samples before the generation phase, enabling a practical defense mechanism for RAG systems operating in the black-box setting.

\begin{table*}[t]
\centering
\caption{Comparison of representative defenses against knowledge poisoning attacks in retrieval-augmented generation systems.}
\label{tab:intro_comparison}
\begin{tabular}{lccccc}
\hline
\textbf{Method} & \textbf{White-box Access Required} & \textbf{Uses LLM Activations} & \textbf{Uses Ensemble Generation} & \textbf{Statistical Detection} & \textbf{Retrieval-Aware} \\
\hline
Tan et al.~\cite{tan2024knowledge} & Yes & Yes & No & No & No \\
Xiang et al.~\cite{xiang2024certifiably} & No & No & Yes & No & No \\
Zou et al.~\cite{zou2024poisonedrag} & No & No & No & Partial & No \\
FilterRAG (Ours) & No & No & No & Yes & Yes \\
ML-FilterRAG (Ours) & No & No & No & Yes & Yes \\
\hline
\end{tabular}
\end{table*}

\section{Conclusion}\label{sec:conclusion}
In this work, we propose a statistical property and establish that the clean and adversarial texts in PoisonedRAG exhibit distinct values for our proposed property. We then propose two defense methods: FilterRAG and ML-FilterRAG that employ the above insight to defend against PoisonedRAG when retrieving texts from RAG's knowledge database. Our methods demonstrate robustness under various LLMs. Experimental results show that our proposed methods achieve tremendous performance with a performance gap difference of merely up to 0.2\% compared to the original RAG. In general, our methods can defend robustly against PoisonedRAG in the black-box setting.

\textbf{Limitations and Future Works:}
The following are the limitations of our work.
\begin{itemize}
    \item Our work focuses primarily on addressing the PoisonedRAG knowledge poisoning attack. It is not tested against other forms of RAG knowledge poisoning attacks such as emotional-based and topic-flipping attacks. Future explorations are required to investigate how it works against the mentioned attacks. 
    \item For the ML-FilterRAG method, we only use simple machine learning models to detect adversarial texts. In the future, the use of more extensive machine learning models can be explored. 
    \item Our proposed methods add a component (filtration) to the traditional RAG. This introduces an additional computation demand. The analysis of this computational requirement and its solutions can be explored in the future. 
\end{itemize}

\begin{table*}[htbp]
    \centering
    \caption{Performance Comparison of RAG Frameworks Across Different LLMs for White-box Attack}
    \label{tab:main-performance-white-box}
    \begin{tabular}{@{}ll c cc cc cc cc@{}} 
        \toprule 
        \textbf{Dataset} & \textbf{Framework} & \textbf{ATR} $\downarrow$& \multicolumn{2}{c}{\textbf{GPT-3.5}} & \multicolumn{2}{c}{\textbf{GPT-4}} & \multicolumn{2}{c}{\textbf{GPT-4o}} & \multicolumn{2}{c}{\textbf{LLaMA-3}}\\
        \cmidrule(lr){4-5} \cmidrule(lr){6-7} \cmidrule(lr){8-9} \cmidrule(lr){10-11}
        & & & \textbf{ASR} $\downarrow$ & \textbf{Accuracy} $\uparrow$& \textbf{ASR} $\downarrow$ & \textbf{Accuracy} $\uparrow$& \textbf{ASR} $\downarrow$ & \textbf{Accuracy} $\uparrow$ &  \textbf{ASR} $\downarrow$ & \textbf{Accuracy} $\uparrow$\\
        \midrule 
        \multirow{4}{*}{HotpotQA} & CleanRAG & \textbf{0.000} & 0.080 & \textbf{0.913} & \textbf{0.090}& \textbf{0.908}& 0.090& \textbf{0.907} & \textbf{0.310} & \textbf{0.665}\\
        & PoisonedRAG & 1.000 & 0.860 & 0.056 & 0.870& 0.022 &0.872 & 0.021 &0.980 & 0.001\\
        & FilterRAG ($\epsilon=0.2$) & \textbf{0.000} & \textbf{0.005} & 0.721 & \textbf{0.090}&0.733 &0.090 & 0.731 & 0.335 & 0.600\\
        & ML-FilterRAG & 0.030 & 0.090 & 0.729 &\textbf{0.090} &0.745 &\textbf{0.080} & 0.755 & 0.330 & 0.650\\
        
        \midrule 
        \multirow{4}{*}{MS-MARCO} & CleanRAG & \textbf{0.000} & \textbf{0.060} & \textbf{0.859} & \textbf{0.050}& \textbf{0.851} & \textbf{0.050} & \textbf{0.867} & \textbf{0.310} & \textbf{0.667}\\
        & PoisonedRAG & 0.980 & 0.720 & 0.074 & 0.710 & 0.107 & 0.750 & 0.102 & 0.960 &0.030\\
        & FilterRAG ($\epsilon=0.2$) & 0.125 & 0.120 & 0.721 & 0.110 & 0.729 & 0.110 & 0.706 & 0.320 & 0.596\\
        & ML-FilterRAG & 0.155 & 0.110 & 0.762 & 0.110 & 0.777 & 0.112 & 0.782 & 0.318 & 0.629\\
        
        \midrule 
        \multirow{4}{*}{NQ} & CleanRAG & \textbf{0.000} & \textbf{0.010} & \textbf{0.831} & \textbf{0.020} & \textbf{0.833} & \textbf{0.020} & \textbf{0.831} &\textbf{0.140} & \textbf{0.575}\\
        & PoisonedRAG & 1.000 & 0.730 & 0.031 & 0.700 & 0.033 & 0.720 & 0.046 & 0.970 & 0.010 \\
        & FilterRAG ($\epsilon=0.2$) & 0.025 & 0.050 & 0.823 & 0.050 & 0.816 & 0.050 & 0.828 & 0.160 & 0.500\\
        & ML-FilterRAG & 0.040& 0.060 & 0.824 & 0.050 & 0.827 & 0.050 & 0.829 & 0.170 & 0.542\\
        \bottomrule 
    \end{tabular}
\end{table*}

\ifCLASSOPTIONcaptionsoff
  \newpage
\fi

\bibliographystyle{IEEEtran}
\bibliography{ref}

\appendices
\section{Performance for White-box Attacks}\label{app:white-box}
Further results of our proposed methods for white-box attack are shown in Table \ref{tab:main-performance-white-box}. From the results, we observe that the results show pattern similar to that of black-box attack.

\section{Prompt Template}\label{app:prompt}
\begin{figure}[htbp]
    \centering
    \includegraphics[width=\columnwidth]{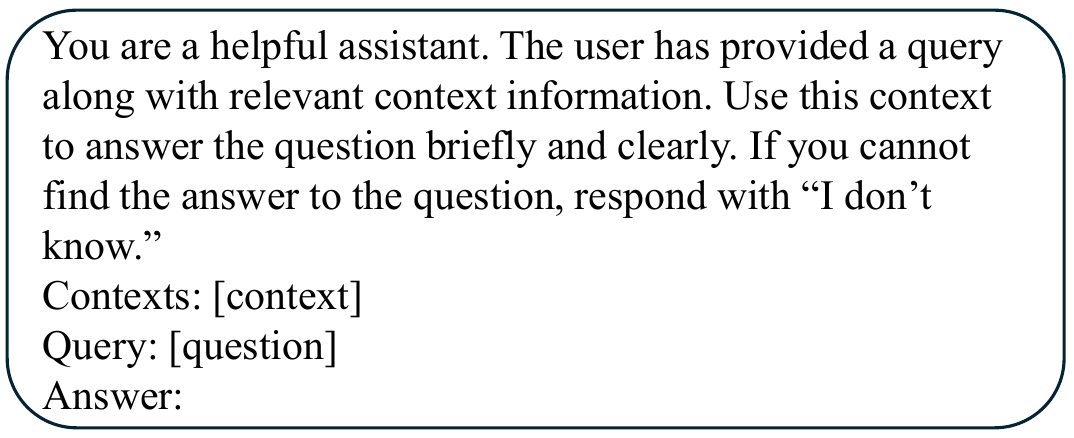} 
    \caption{The prompt template used to produce the results \cite{zou2024poisonedrag}} 
    \label{fig:prompt}
\end{figure}

Figure \ref{fig:prompt} shows the prompt template used in our experiments. We adopt the same prompt template used in \cite{zou2024poisonedrag} in our experiments. 

\section{ML-FilterRAG Machine Learning Models}\label{app:training-details}
Table \ref{tab:training-details} presents the performance of the machine learning models trained for ML-FilterRAG. We present the training and test accuracies for the different machine learning models trained for the different datasets used in our experiments. The accuracy values show that the machine model trained on MS-MARCO dataset performs worse compared to the model trained on NQ and HotpotQA datasets. 

\begin{table}[htbp]
    \centering
    \caption{ML-FilterRAG Model Training Results}
    \label{tab:training-details}
    \begin{tabular}{@{}llcc@{}} 
        \toprule 
         \textbf{Dataset}&\textbf{Model} & \textbf{Training Accuracy} & \textbf{Test Accuracy}\\
         \midrule 
         MS-MARCO & Random Forest& 0.860& 0.850\\
         NQ &  XGBoost& 0.990 & 0.970\\
         HotpotQA & Random Forest& 0.990& 0.982\\
        
        \bottomrule 
    \end{tabular}
\end{table}

\newpage

\section{Feature Pair Plots} \label{app:pairplots}
\begin{figure}[htp]
    \centering
    \includegraphics[width=\columnwidth]{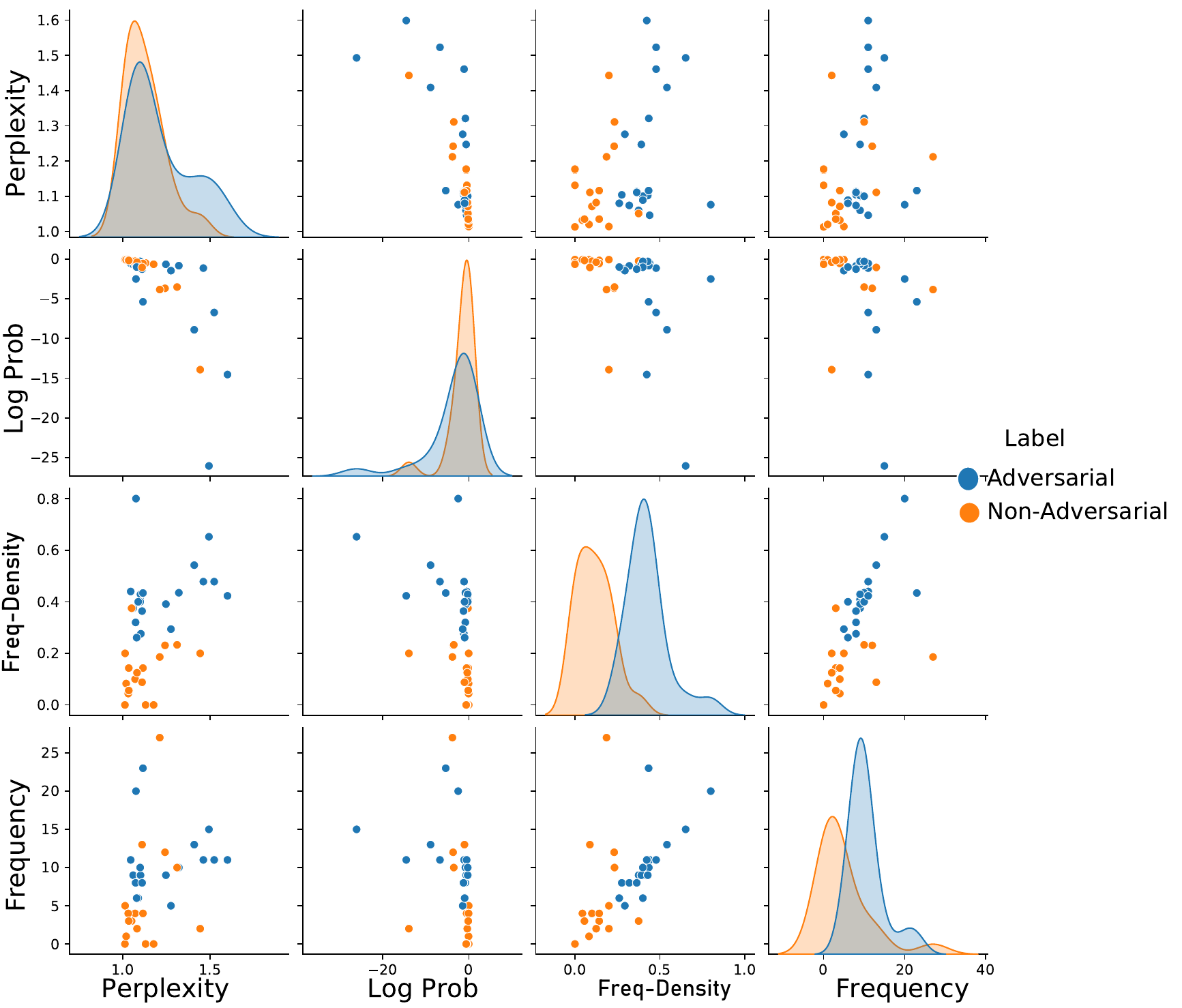} 
    \caption{Pair plots for all the features} 
    \label{fig:pairplots}
\end{figure}

Figure \ref{fig:pairplots} shows the detailed feature pair-plots used to filter out adversarial texts in this work. 

\end{document}